\documentclass[letterpaper, 10 pt, conference]{ieeeconf}
\usepackage[utf8]{inputenc} 

\IEEEoverridecommandlockouts                              

\usepackage{graphics}
\usepackage{epsfig}
\usepackage{graphicx}
\usepackage{grffile}

\usepackage{mathptmx}
\usepackage{times}

\usepackage{amsmath}
\usepackage{amssymb}

\usepackage{array}
\usepackage{booktabs}
\usepackage{tabularx}
\newcolumntype{Y}{>{\centering\arraybackslash}X}
\usepackage{multirow}
\usepackage{makecell}
\usepackage{ragged2e}   

\usepackage[table]{xcolor}
\definecolor{oursgray}{RGB}{235,235,235}
\newcommand{\oc}[1]{\cellcolor{oursgray}#1}

\usepackage{caption}
\usepackage{subcaption}
\usepackage{capt-of}    

\usepackage{cuted}      

\usepackage{eso-pic}

\usepackage{url}
\usepackage[hidelinks]{hyperref}

\usepackage{pifont}
\newcommand{\cmark}{\textcolor{green!60!black}{\ding{51}}}  
\newcommand{\xmark}{\textcolor{red!70!black}{\ding{55}}}    

\usepackage{placeins}   
\usepackage{balance}    

\title{\LARGE \bf
 NeuroCommitSSM: Decision-Centric Shared Autonomy for Safe Assistive Manipulation via EEG--EMG--ET Commit Readiness
}

\author{%
Tipu~Sultan$^{1}$,
Param~Sangani$^{2}$,
Kody~Cool$^{1}$,
Pascal~Sikorski$^{2}$,
Guangping~Liu$^{1}$,\\
Hadi~Akbarpour$^{2}$,
and Madi~Babaiasl$^{1}$%
\thanks{$^{1}$Aerospace and Mechanical Engineering Department, Saint Louis University, St. Louis, MO, USA.
Email: {\tt\small tipu.sultan@slu.edu, madi.babaiasl@slu.edu.}
$^{2}$Computer Science Department, Saint Louis University, St. Louis, MO, USA.
*This work has been supported by SLU's Research Institute under Award-01935, Clare Boothe Luce Foundation under Award-01281, and SLU's AEME department under Proj-000480.}%
}

\newcommand{\ArxivIEEENotice}{%
  \AddToShipoutPictureFG*{%
    \AtPageLowerLeft{%
      \put(\LenToUnit{0.62in},\LenToUnit{0.20in}){%
        \parbox{7.25in}{%
          \centering
          \fontsize{5.2}{5.8}\selectfont
          \textit{Accepted at the 2026 IEEE/RSJ International Conference on Intelligent Robots and Systems (IROS 2026).}\\[-0.15ex]
          \textcopyright{} 2026 IEEE. Personal use of this material is permitted. Permission from IEEE must be obtained for all other uses, in any current or future media, including reprinting/republishing this material for advertising or promotional purposes, creating new collective works, for resale or redistribution to servers or lists, or reuse of any copyrighted component of this work in other works.%
        }%
      }%
    }%
  }%
}

\begin{document}
\raggedbottom

\maketitle
\ArxivIEEENotice
\thispagestyle{empty}
\pagestyle{empty}

\begin{strip}
\centering
\vspace{-7.0em}
\includegraphics[width=\textwidth,keepaspectratio]{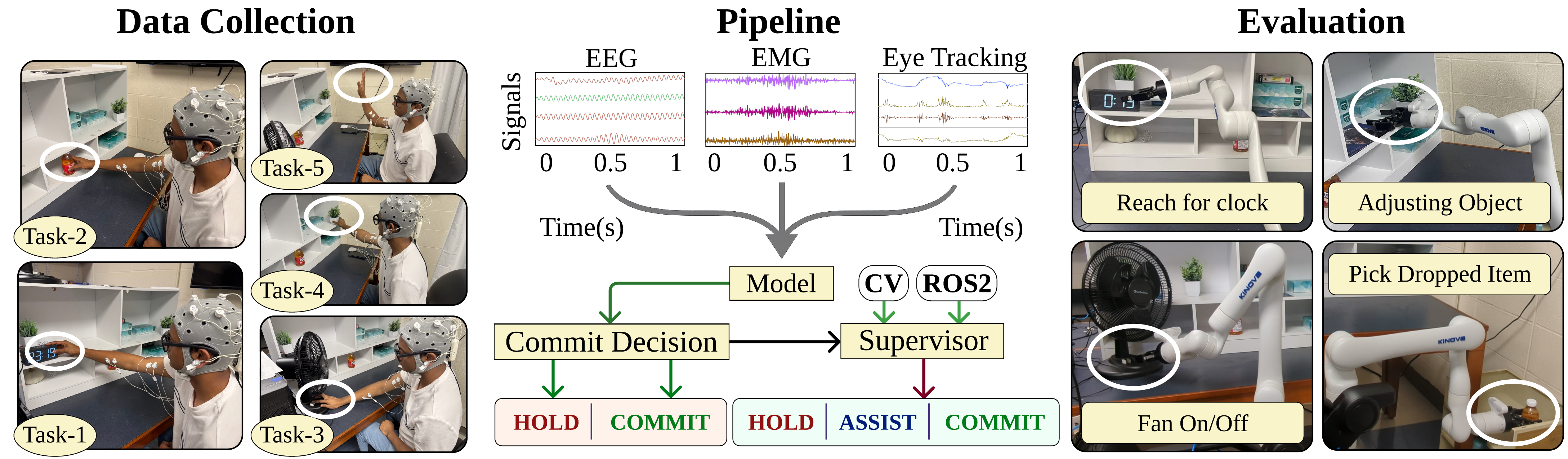}

{\setlength{\abovecaptionskip}{2pt}\setlength{\belowcaptionskip}{-6pt}

\captionof{figure}{NeuroCommitSSM overview. \textbf{Left:} Synchronized electroencephalogram (EEG), electromyography (EMG), and eye-tracking (ET) signals from five International Classification of Functioning, Disability and Health (ICF)-aligned assistive tasks. \textbf{Middle:} Time-aligned windows yield a continuous commit-readiness score and a HOLD/COMMIT decision, while a Robot Operating System (ROS2)/computer vision (CV) supervisor enforces action feasibility via HOLD--ASSIST--COMMIT shared autonomy. \textbf{Right:} Kinova Gen3 evaluations on five Activities of Daily Living (ADL) tasks (e.g., clock reach, object adjustment, fan on/off, item pickup).}

\label{fig:teaser}}

\vspace{-1.0em}
\end{strip}


\begin{abstract}
We present NeuroCommitSSM, a decision-centric framework that models \emph{when} to execute, not just \emph{what} to do, for safe commit-to-execute control in assistive robotic manipulation. To address this gap, NeuroCommitSSM predicts a continuous commit-readiness score $c_t\!\in\![0,1]$ from synchronized electroencephalography (EEG), electromyography (EMG), and eye-tracking (ET), and converts it into discrete commit events via dwell/hysteresis filtering. A three-state finite-state supervisor (HOLD--ASSIST--COMMIT (HAC)) gates execution by requiring both a sustained commit-readiness signal from the neural model \emph{and} real-time perception and robot-state feasibility (target visibility, Inverse Kinematics (IK) solvability, collision-free planning) to be simultaneously satisfied before initiating motion. We evaluate on $N{=}32$ subjects performing five activities of daily living (ADL) tasks aligned with the International Classification of Functioning, Disability and Health (ICF), under leave-one-subject-out (LOSO) cross-validation and seven sensor-dropout scenarios (S0--S6). NeuroCommitSSM achieves 0.950 action-balanced accuracy with only 0.75 false commit events per 1000 REST windows (FP/1k REST), and maintains low false commits and stable state transitions under sensor loss (e.g., EEG-only: 0.785 balanced accuracy, 0.29 FP/1k REST), while baselines collapse under the same conditions (e.g., Temporal Convolutional Network (TCN): 99.95 FP/1k REST). Hardware-in-the-loop (HIL) validation on a Kinova Gen3 arm shows that feasibility-checked execution reduces false starts and decision instability without sacrificing task success. Supplementary materials (codes/datasets/videos/extra analyses) are available at \href{https://madibabaiasl.github.io/NeuroCommitSSM/}{https://madibabaiasl.github.io/NeuroCommitSSM/}.
\end{abstract}








\section{Introduction}

Autonomous assistive robotic manipulators can restore independence in daily living activities by mapping low-dimensional, hands-free control interfaces to arm functions within a shared-control framework. This enables users to express intent while the arm assists with tasks such as opening door handles, eating, and drinking \cite{hansen2024multimodal}. In real-world daily use and clinical neurorehabilitation, a key challenge is not only to decipher \emph{what} users intend, but understand \emph{when robotic action is safe to execute}. Non-target EMG activity during activities of daily living can falsely activate closed-set control systems, producing erroneous inputs for robotic execution while inducing user frustration. Therefore, false activations during \textsc{Rest} (periods when the user has no intent to act) should be minimized \cite{eddy2025emg}. This timing safety gap is particularly severe for multimodal biosignals (e.g., EEG/EMG), which are noisy, nonstationary, and highly subject-dependent. This motivates decision-centric interfaces that emphasize stable, low-false-positive \emph{commit} decisions (not just window-level intent scores) and leverage additional modalities, such as gaze/eye-tracking, to improve accuracy \cite{dillen2024shared,abdallah2025hybrid}.


Most intent-recognition pipelines operate on windows and trigger execution by thresholding a classifier or control score, often stabilizing decisions with temporal smoothing (e.g., exponential moving average (EMA)) or multi-prediction aggregation (e.g., consecutive-detection rules) \cite{hussein2022development,yu2023gesture,eddy2025emg}. Raw intent confidence can be poorly calibrated for safety-critical actuation and unstable near rest--action transitions, motivating hysteresis gating (separate commit-on/off thresholds) and time-based rejection of brief spikes \cite{gaus2026act}. False positives during rest are particularly harmful in rehabilitative motor-imagery brain--computer interfaces (MI-BCI) \cite{jeong2025investigating}, motivating mechanisms to suppress or penalize unintended initiations. Even with correct intent, execution can be infeasible due to obstacles, unreachable poses, or joint limits, requiring confirmation of motion feasibility by validating inverse kinematic solutions and planned avoidance of potential on-path collisions to prevent or refine motion \cite{bustamante2024feasibility}. Without confidence and feasibility safeguards, intent-triggered assistance can be premature and disruptive, causing erroneous assistance and user frustration \cite{liu2025casper}. In deployed systems, feasibility gating depends on perception pipelines that localize task-relevant targets and scene constraints; many systems use fiducial markers (e.g., AprilTag) for target detection and pose estimation \cite{olson2011apriltag}, pre-built object models or template-based pose pipelines \cite{hinterstoisser2012model}, or modern learned RGB-D 6D pose estimators (e.g., FoundationPose) \cite{wen2024foundationpose}, whereas our implementation uses a markerless RGB-D geometric pipeline (detection/segmentation, depth backprojection, and planar contact-region estimation) with object fitting via RANSAC-based geometric estimation \cite{lim2024multi}.




This paper makes intent-to-execution \emph{decision-centric} (\textbf{Figure~\ref{fig:teaser}}). NeuroCommitSSM learns a continuous commit-readiness signal $c_t\!\in\![0,1]$ from synchronized EEG--EMG--ET using neurophysiology-aware encoders and entropy-regularized multimodal fusion, and converts it to commit events via dwell/hysteresis filtering. Commit quality is measured under LOSO cross-subject shift and sensor dropout scenarios (S0--S6; S0: all sensors, S1--S3: one sensor missing, S4--S6: unimodal) using FP/1k REST and flaps/min (state toggles per minute). Our HAC supervisor integrates commit readiness with perception and robot-state feasibility checks, validated via HIL biosignal replay on a Kinova Gen3 arm. These considerations motivate the following research question: \textbf{RQ:} \emph{Does commit-readiness with dwell/hysteresis filtering and feasibility gating reduce \textsc{Rest} false activations and state flapping while preserving task success and responsiveness under LOSO shift and sensor dropout (S0--S6)?} Despite rapid advances in intent decoding, synchronized datasets for \emph{functionally complete} assistive ADL skills remain absent. We release (to our knowledge) the first synchronized EEG--EMG--ET dataset for ICF-aligned ADL intent decoding benchmarked under balanced-window LOSO. Our contributions are:



\begin{itemize}
    \item NeuroCommitSSM, a decision-centric intent-to-commit model that predicts a continuous commit score $c_t$ from synchronized EEG--EMG--ET and converts it to commit events using dwell/hysteresis filtering, with entropy-regularized multimodal fusion under sensor dropout (evaluated under LOSO using FP/1k \textsc{Rest} and flaps/min).

    \item A ROS2-native HAC shared-autonomy framework that integrates commit-readiness gating with a task-aware, markerless RGB-D perception pipeline to provide scene/robot feasibility checks for safe execution, including anti-flap safeguards and abort/recovery behavior, without fiducial markers, CAD model registration, or task-specific learned 6D pose estimators.
    
    
    
    \item System-level validation on a real Kinova Gen3 arm using HIL replay of recorded biosignals, demonstrating execution success and safety through unsafe-motion prevention and feasibility-triggered abort analysis.

    \item To the best of our knowledge, we release the first synchronized EEG--EMG--ET dataset for \emph{functionally complete}, ICF-aligned assistive ADL intent decoding, benchmarked under a balanced-window LOSO protocol.
\end{itemize}

\begin{table*}[!htbp]
\centering
\caption{Comparison of biosignal datasets and evaluation protocols for assistive intent decoding, highlighting modality coverage, ADL relevance, and whether evaluation is limited to offline open-loop benchmarks or includes system-level closed-loop/robot-in-the-loop validation (ours: HIL replay through the full robot stack).}

\label{tab:dataset_compare}
\setlength{\tabcolsep}{4.0pt}
\renewcommand{\arraystretch}{1.12}
\scriptsize

\begin{tabularx}{\textwidth}{@{}l c c c c c c c c Y@{}}
\toprule
\multirow{2}{*}{\textbf{Dataset}} &
\multicolumn{3}{c}{\textbf{Context}} &
\multicolumn{3}{c}{\textbf{Data modality}} &
\multicolumn{2}{c}{\textbf{Protocol}} &
\multirow{2}{*}{\textbf{Task set / notes}} \\
\cmidrule(lr){2-4}\cmidrule(lr){5-7}\cmidrule(lr){8-9}
& \textbf{Domain} & \textbf{Pop.} & \textbf{Setting}
& \textbf{EEG} & \textbf{EMG} & \textbf{Eye}
& \textbf{ADL skills} & \textbf{Closed-loop} & \\
\midrule

\cite{dere2022_emg_eeg_gesture_mendeley} &
Gesture/rehab & N=11 & Lab &
\cmark & \cmark & \xmark &
\xmark & \xmark &
\makecell[c]{7 gestures, MI+execution, cue-based rest/hold.} \\

\cite{xi2019_eeg_emg_coherence_mendeley} &
Coherence & N=5 & Lab &
\cmark & \cmark & \xmark &
\xmark & \xmark &
\makecell[c]{4 wrist/fist actions, offline EEG--EMG coherence.} \\

\cite{garro2025eeg} &
Reaching & N=40 & Lab &
\cmark & \cmark & \xmark &
\xmark & \xmark &
\makecell[c]{3 reaches, 10 repetitions each, exoskeleton-assisted.} \\

\cite{guo2021motor_imagery_data} &
Motor imagery & N=8 & Lab &
\cmark & \cmark & \xmark &
\xmark & \xmark &
\makecell[c]{3 MI classes, wrist flexion, elbow flexion, fist clench.} \\

\cite{pizzolato2017comparison} &
sEMG gestures & N=10 & Lab &
\xmark & \cmark & \xmark &
\xmark & \xmark &
\makecell[c]{52 movements + rest, 6 repetitions per movement.} \\

\cite{cognolato2020gaze} &
Prosthetics/Grasping & N=45 & Lab &
\xmark & \cmark & \cmark &
\cmark & \xmark &
\makecell[c]{10 household grasps, gaze + EMG prosthetic control.} \\

\midrule
\textbf{Ours (ICF-ADL)} &
\textbf{Assistive ADL} & \textbf{N=32} & \makecell[c]{\textbf{Lab}\\\textbf{Free env.}} &
\textbf{\cmark} & \textbf{\cmark} & \textbf{\cmark} &
\textbf{\cmark} & \textbf{\cmark} &
\makecell[c]{\textbf{Functionally complete, ICF-aligned skills,}\\ \textbf{balanced-window LOSO benchmark.}} \\

\bottomrule
\end{tabularx}
\end{table*}


\section{Related Work}
\label{sec:related_work}

Shared autonomy for biosignal-driven assistive arms reduces the burden of high-degree-of-freedom (DoF) control by inferring user intent from physiological signals and arbitrating execution between user and robot \cite{ douglas2025levels}. Prior systems emphasize intent or goal inference, switching control when uncertain, and uncertainty-aware arbitration to improve efficiency and usability in ADL-like manipulation \cite{tao2024incremental,jonnavittula2024sari}. However, pipelines treat the \emph{trigger} from intent estimates to robot motion as an empirical step rather than a first-class decision problem.


Eye-tracking provides fast, task-relevant cues, but is sensitive to calibration drift, occlusions, and noise, motivating system reliability against degraded gaze streams \cite{hyung2025robotic,belardinelli2022intention,
fischer2024scoping}. Shared-autonomy systems also use perception-derived world models and robot-state/motion feasibility checks to ensure that actions are executable \cite{bustamante2024feasibility}, but they typically do not formalize a multi-state supervisor that couples readiness with feasibility. Feasibility answers whether the robot \emph{can} act, not whether the user is ready to commit. Unlike marker- or model-based pose pipelines \cite{olson2011apriltag,hinterstoisser2012model,wen2024foundationpose}, existing perception modules are not integrated with decision-centric commit gating for closed-loop assistive execution.



BCI and physiological decoding research has explored multimodal fusion of EEG, EMG, and ET signals to improve intent recognition performance 
\cite{abdallah2025hybrid,yi2024hybrid}. Although these methods improve classification accuracy, they focus on window-level decoding and typically trigger actions by thresholding confidence scores, leaving unintended initiations and decision instability in continuous streaming settings insufficiently addressed. False activations at rest and miscalibrated confidence-based triggering remain major barriers, especially near transitions such as gaze shifts \cite{guo2023robotic,eddy2025emg,miladinovic2024optimizing}. Common mitigations (e.g., EMA, majority vote, hysteresis) improve smoothness, but they still conflate recognition confidence with readiness to commit. They do not model commit readiness as a temporal decision objective with false-activation penalties, nor do they integrate intent readiness with robot-state feasibility in a unified supervisor.


Most publicly available biosignal datasets for intent recognition target \emph{open-loop} offline decoding of simplified primitives, such as gestures, motor imagery, coherence analysis, or constrained reaching collected in controlled laboratory settings \cite{dere2022_emg_eeg_gesture_mendeley,xi2019_eeg_emg_coherence_mendeley,garro2025eeg,guo2021motor_imagery_data,pizzolato2017comparison}. ET is beginning to appear in assistive contexts; for example, MeganePro pairs gaze with EMG for household grasping, but does not include synchronized EEG and remains open-loop \cite{cognolato2020gaze}. As summarized in \textbf{Table~\ref{tab:dataset_compare}}, prior resources rarely provide a fully \emph{synchronized} EEG--EMG--ET combination together with \emph{ADL-style} tasks and a standardized benchmark protocol suitable for cross-subject comparison \cite{dere2022_emg_eeg_gesture_mendeley,xi2019_eeg_emg_coherence_mendeley,garro2025eeg,guo2021motor_imagery_data,pizzolato2017comparison,cognolato2020gaze}. To overcome this limitation, we release a synchronized EEG--EMG--ET dataset for functionally complete, ICF-aligned assistive ADL intent decoding and provide a reproducible balanced-window LOSO benchmark, with system-level evaluation via HIL replay on a Kinova Gen3 arm.


\begin{figure*}[!t]
    \centering
    \includegraphics[width=\linewidth]{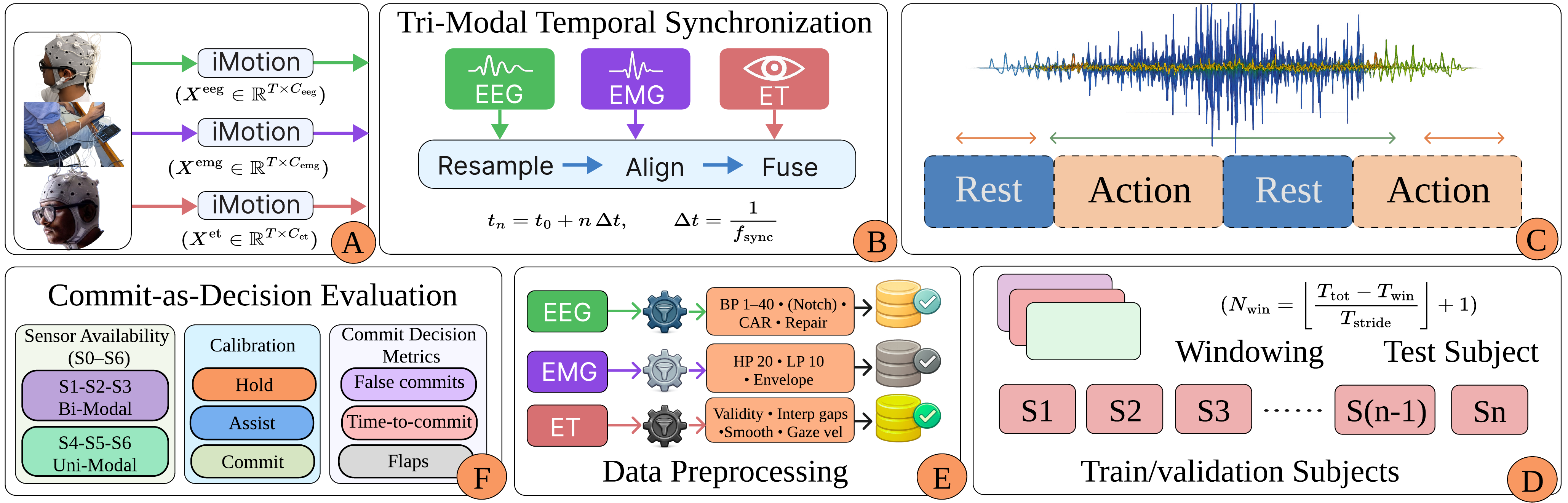}
\caption{Overview of the intent-to-commit pipeline: (A) windowed EEG/EMG/ET with masks (acquired through iMotion software); (B) tri-modal resampling and alignment; (C) REST/ACTION segmentation with onsets; (D) sliding windows and LOSO; (E) deterministic preprocessing; (F) commit-as-decision evaluation under sensor dropout S0--S6 with HAC calibration. BP, HP, and LP = Band-pass, High-pass, and Low-pass filters, CAR = common average referencing, Interp. = Interpretation, vel. = velocity. }
    \label{figure:biorob_pipeline}
\end{figure*}


\section{Methodology}
\label{sec:method}

An end-to-end \emph{intent-to-commit} pipeline converts synchronized EEG, EMG, and ET streams into windows, per-window intent predictions, and \emph{commit-to-execute} decisions for shared autonomy (\textbf{Figure~\ref{figure:biorob_pipeline}}). Time-aligned streams are preprocessed to produce modality-specific sequences and validity masks, then segmented into \textsc{Rest} and \textsc{Action} regions using a consensus Hidden Semi-Markov Model (HSMM)-based labeling pipeline. Given a tri-modal window, NeuroCommitSSM jointly predicts (i) \textsc{Rest} vs.\ \textsc{Action}, (ii) task class (T1--T5) for \textsc{Action} windows, and (iii) continuous readiness score $c\in[0,1]$. A dwell/hysteresis filter converts $c$ into discrete events, passed to a shared-autonomy supervisor (HOLD/ASSIST/COMMIT) with robot feasibility checks to gate safe execution. Performance is evaluated under nominal sensing and sensor dropout using LOSO splits for subject-independent generalization.

\subsection{Dataset and Experimental Protocol}
\label{subsec:dataset_protocol}

A synchronized tri-modal dataset was collected from $N{=}32$ adults in a laboratory setting that simulated a lifelike scenario involving a chair, table, and shelf. Participant demographics and technology familiarity were recorded via questionnaire (age, gender, prior assistive-technology use, and comfort with new technology). Participants were $25.03 \pm 6.03$ years old (mean $\pm$ SD; range: 18--46), with gender counts of $M{=}22$ and $F{=}10$. Most reported no prior assistive-technology experience ($n{=}29$; prior use $n{=}3$). Each session lasted 90--120 minutes (sensor setup, calibration, task execution, and breaks), totaling 64 hours across the study. The protocol was approved by the University IRB, and all participants provided written informed consent. Six intent classes were defined: \textsc{Rest} and five ICF-aligned ADL tasks: T1 (\textit{Clock}): grasp tabletop clock; T2 (\textit{Bottle}): retrieve dropped bottle; T3 (\textit{Fan}): toggle fan on/off; T4 (\textit{Plant}): reposition shelf plant; and T5 (\textit{Wave}): greeting gesture. Each subject performed 16 repetitions per task, yielding 2656 total trials. Thirty subjects were used for LOSO model development, 2 were held out for robot evaluation. EEG, EMG, and ET were recorded in iMotions via Lab Streaming Layer (LSL). EEG was acquired with an Enobio-8 (8 channels, 250\,Hz; F3, C3, C4, P3, Fz, Cz, Pz, Oz). Surface EMG was recorded with a 4-channel PLUX system over upper-limb muscles (biceps, triceps, deltoid, forearm flexors) following SENIAM guidelines \cite{hermens2000development}. ET was captured with Pupil Labs Neon glasses. An HSMM-based labeling pipeline generated \textsc{Action}/\textsc{Rest} supervision labels from EEG, EMG, eye-tracking activity features, and trial-timing constraints. A single-action HSMM with duration priors and rest-region constraints selected the final \textsc{Action} segment, which was refined using EMG boundary and quiet-window checks. Per-sample labels were converted to window labels by majority overlap with the resulting segmentation.


\subsection{Signal Preprocessing and Windowing}
\label{subsec:preproc}

A deterministic pipeline converts synchronized streams into masked, model-ready windows. Signals are resampled to 250\,Hz when the native rate deviates by more than 1\%. Fixed-length windows of $T{=}500$ samples (2.0\,s) are extracted with a 0.25\,s stride and labeled by majority vote. EEG is band-pass filtered (1--40\,Hz), auto-notch filtered at 50/60\,Hz, re-referenced via common-average referencing (CAR), and repaired with Hampel-based outlier correction. EMG is high-pass filtered (20\,Hz), rectified, and low-pass filtered (10\,Hz) for amplitude envelope extraction. ET is cleaned via validity/worn-state checks, short-gap interpolation (${\le}150$\,ms), and temporal smoothing. Fold-wise normalization applies training-subject statistics to validation/test splits. Compact engineered descriptors are concatenated with learned representations to improve reliability under sensor dropout. EEG descriptors include PSD/bandpower summaries computed from the preprocessed EEG window across standard frequency bands. EMG descriptors include root mean square (RMS), mean absolute value (MAV), and waveform length (WL) computed per EMG channel over each 2.0-s window.


 \subsection{Problem Formulation}
Given a synchronized tri-modal window, NeuroCommitSSM predicts (i) action (\textsc{Rest} vs.\ \textsc{Action}), (ii) task class (T1--T5), and (iii) commit-readiness score $c\in[0,1]$, converted to discrete commit events via a dwell/hysteresis filter for shared-autonomy execution.
\subsection{Tri-modal Inputs and Availability Scenarios}
Each window (of length $T$ samples) provides tensors $\mathbf{X}^e\!\in\!\mathbb{R}^{T\times 8}$, $\mathbf{X}^m\!\in\!\mathbb{R}^{T\times 4}$, and $\mathbf{X}^{et}\!\in\!\mathbb{R}^{T\times C_{et}}$, where $C_{et}$ denotes the number of eye-tracking feature channels. Per-sample validity masks $\mathbf{M}^e\!\in\![0,1]^{T\times 8}$, $\mathbf{M}^m\!\in\![0,1]^{T\times 4}$, and $\mathbf{M}^{et}\!\in\![0,1]^{T\times C_{et}}$ are propagated through tokenization, fusion, and pooling to suppress invalid contributions. Per-modality reliability cues $r^e,r^m,r^{et}\!\in\!\mathbb{R}^{d_r}$ ($d_r{=}4$) are computed from lightweight quality proxies (coverage fraction, missingness, flatline/clip indicators). An auxiliary feature vector $\mathbf{F}=[F_{psd},F_{emg},F_{mask}]$ provides EEG spectral, EMG time-domain, and modality-availability summaries when enabled. Performance is evaluated over seven sensor dropout scenarios (EEG, EMG, ET): S0$(1,1,1)$, S1$(0,1,1)$, S2$(1,0,1)$, S3$(1,1,0)$, S4$(1,0,0)$, S5$(0,1,0)$, S6$(0,0,1)$. 


 \subsubsection{NeuroCommitSSM Architecture}
\begin{figure*}[t]
    \centering
    \includegraphics[width=\linewidth]{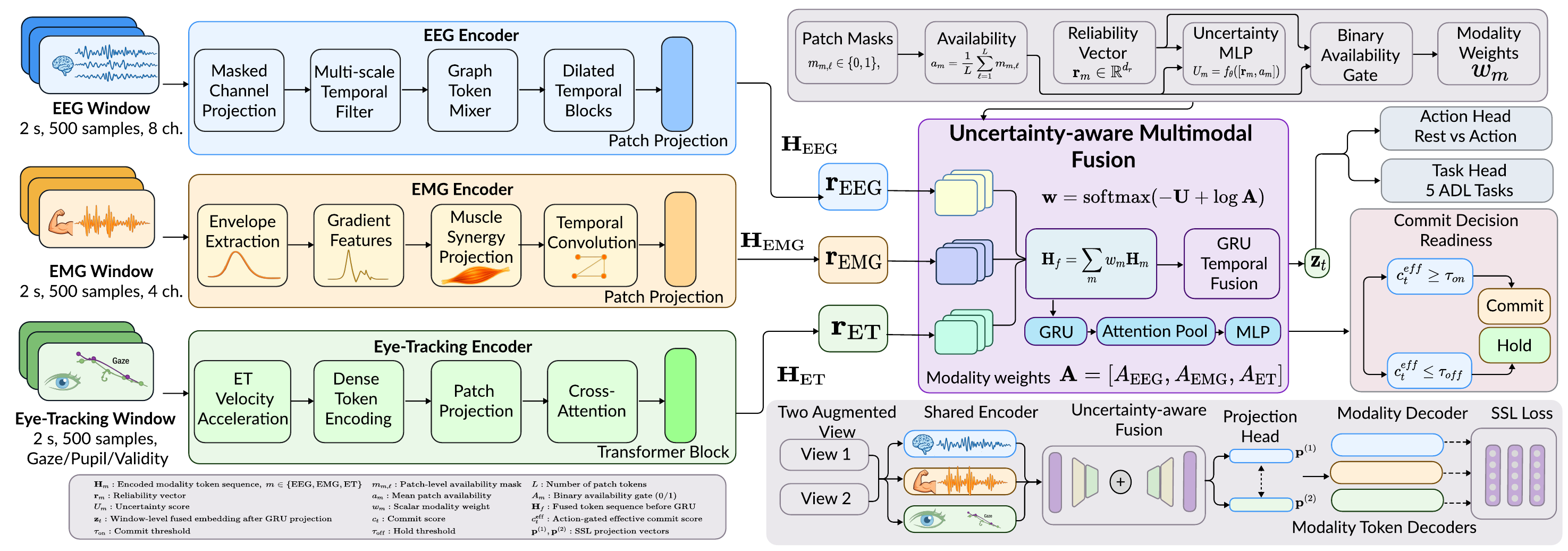}
    \caption{NeuroCommitSSM architecture. Masked EEG/EMG/ET windows are tokenized and encoded by modality-specific temporal encoders, then fused via reliability-aware token-level integration. Action/task heads output class probabilities, and a commit head predicts a continuous readiness score $c_t\!\in\![0,1]$ used for dwell/hysteresis event filtering.}

    \label{figure:architecture}
\end{figure*}

\textbf{Figure~\ref{figure:architecture}} shows the NeuroCommitSSM architecture. The model tokenizes each modality into patch sequences, fuses them with uncertainty-aware weighting, and predicts action, task, and commit readiness. Window length $T{=}500$ (2.0\,s at 250\,Hz), patch size $p{=}25$, embedding dimension $D{=}160$, and token length $L{=}20$. Given input $\mathbf{V}\!\in\!\mathbb{R}^{B\times T\times C_v}$ and mask $\mathbf{M}$, elementwise masking and a strided 1D convolution produce patch tokens
$\mathbf{H}=\mathrm{LN}\!\big(\mathrm{Conv1D}_{k=p,s=p}(\mathrm{Drop}((\mathbf{V}\odot\mathbf{M})^\top))^\top\big)\in\mathbb{R}^{B\times L\times D}$.
Patch validity is obtained by pooling and thresholding with $\eta{=}0.10$. The EEG encoder produces masked per-channel features
$\mathbf{f}\!\in\!\mathbb{R}^{B\times T\times 8}$ ($B$: batch size, $T$: timesteps,
$8$: EEG channels) via a multiscale depthwise filterbank. Per-channel descriptors
$\mathbf{d}_c\!\in\!\mathbb{R}^{6}$ (mean, standard deviation, energy, slope, absolute difference,
availability for channel $c$) are mapped via a Multi-layer Perceptron (MLP) to $K{=}6$ virtual channel weights
$W_{k,c}=\mathrm{softmax}_c(\alpha_{k,c})$, where $\alpha_{k,c}$ is the raw MLP
logit, with availability bias $\log(\mathrm{avail}_c{+}\epsilon)$, and $\epsilon$ is a
small constant for numerical stability. Here,
$\mathrm{avail}_c=\frac{1}{T}\sum_t m_{t,c}$ is the fraction of valid samples for
channel $c$. Virtual signals $V^e_{t,k}=\sum_c f_{t,c}W_{k,c}$ and soft masks are
refined by a token-graph mixer and dilated depthwise Gated Linear Unit (GLU) blocks,
yielding EEG patch tokens $(\mathbf{H}^e,\mathbf{m}^e)$ (token embeddings and
patch-validity mask). The EMG encoder transforms masked signals into an envelope
$\mathrm{env}=\mathrm{DWConv}(|x|)$ and a burst proxy
$\mathrm{burst}=\mathrm{DWConv}(|\Delta x|)$, where $\mathrm{DWConv}$ denotes
depthwise convolution and $\Delta x$ denotes the first temporal difference. Mixture $x_{\text{mix}}=0.9\,\mathrm{env}+0.6\,\mathrm{burst}+0.2\,x$ is projected into
$M{=}4$ nonnegative synergy channels via a learned softplus basis
$\mathbf{W}\in\mathbb{R}^{M\times C_m}$ ($C_m{=}4$ EMG channels), yielding EMG patch
tokens $(\mathbf{H}^m,\mathbf{m}^m)$. The ET encoder computes per-timestep validity
$v_{b,t}=\frac{1}{C_{et}}\sum_j m_{b,t,j}$ and velocity/acceleration proxies from
first/second differences of $\mathbf{X}^{et}$. Top-$k{=}3$ salient event tokens (scored
by $e_{b,t}=(0.7\,\mathrm{vel}_{b,t}+0.3\,\mathrm{acc}_{b,t})\,v_{b,t}$) are injected
via cross-attention and a gated residual, followed by mask-aware self-attention to
produce ET patch tokens.

 \subsubsection{Uncertainty-Weighted Fusion and Feature Injection}
Modality availability is computed from patch-validity masks as
$a_{j,b}=\frac{1}{L}\sum_{\ell=1}^{L} m^j_{b,\ell}$, where $L$ is the number of patches,
$j\in\{e,m,et\}$ indexes the modality, and $m^j_{b,\ell}$ denotes the patch-validity
mask value for modality $j$, batch sample $b$, and patch $\ell$.
Reliability cues and availability are mapped to uncertainty scores
$u_{j,b}=u_\theta([\mathbf{r}^j_b,a_{j,b}])$ via a shared MLP, where
$[\cdot,\cdot]$ denotes concatenation and $\mathbf{r}^j_b\in\mathbb{R}^{d_r}$ is a per-modality reliability vector (e.g., coverage fraction, missingness, flatline/clip indicators) for modality $j$ in batch sample $b$. Binary gates
$A_{j,b}=\mathbb{I}[a_{j,b}>\delta]$ ($\delta{=}0.05$, minimum availability threshold)
and fusion weights, with $\mathbf{A}_b=[A_{e,b},A_{m,b},A_{{et},b}]$ and $\varepsilon$
a small constant for numerical stability, are defined as
$\mathbf{w}_b=\mathrm{softmax}\!\left(-[u_{e,b},u_{m,b},u_{{et},b}] + \log(\mathbf{A}_b+\varepsilon)\right)$,
which assigns higher weight to modalities with lower uncertainty and sufficient availability;
token-wise fusion is
$\mathbf{H}^f_{b,\ell}=w_{e,b}\mathbf{H}^e_{b,\ell}+w_{m,b}\mathbf{H}^m_{b,\ell}+w_{{et},b}\mathbf{H}^{et}_{b,\ell}$.
The fused token sequence has shape $\mathbf{H}^f\in\mathbb{R}^{B\times L\times D}$.
A fusion Gated Recurrent Unit (GRU) with hidden size 192 processes $\mathbf{H}^f$; the final GRU hidden state
($192$-d) is then projected to the model dimension $D{=}160$ to form the window embedding
$\mathbf{z}\in\mathbb{R}^{B\times D}$.
Auxiliary features $\mathbf{F}_b\in\mathbb{R}^{123}$ combine EEG bandpower summaries $\mathbf{F}_{psd}\in\mathbb{R}^{96}$, EMG time-domain summaries $\mathbf{F}_{emg}\in\mathbb{R}^{24}$, and modality-missingness flags $\mathbf{F}_{mask}\in\mathbb{R}^{3}$ (one per modality). An MLP projects $\mathbf{F}_b$ to $\mathbf{g}_b\in\mathbb{R}^{D}$, which is added to the window embedding as $\mathbf{z}_b\leftarrow\mathbf{z}_b+0.25\,\mathbf{g}_b$. We append $0.25\,\mathbf{g}_b$ as an extra token to $\mathbf{H}^f$, increasing the token length from $L$ to $L{+}1$ and improving performance under sensor dropout.

\subsubsection{Commit-to-Execute Decision, Training, and Calibration}
From $\mathbf{z}$, action logits $\boldsymbol{\ell}^{act}\in\mathbb{R}^{2}$ and task logits $\boldsymbol{\ell}^{task}\in\mathbb{R}^{5}$ are produced; action probability is $p^{act}=\mathrm{softmax}(\boldsymbol{\ell}^{act})_1$. Task supervision applies only to valid \textsc{Action} windows. The commit head processes $(\mathbf{H}^f,m^f)$ via a token-level GRU with mask-aware attention pooling and MLP to output readiness score $c=\sigma(\ell^{ct})\in[0,1]$ and auxiliary state logits $\boldsymbol{\ell}^{state}\in\mathbb{R}^{2}$ (HOLD vs.\ COMMIT). Readiness is supervised from onset-relative timing for \textsc{Action} windows near estimated onset; supervision is disabled for \textsc{Rest} windows. Effective readiness is then action-gated: $\tilde{c}_t=\mathrm{clip}\!\left(c_t\cdot\mathbb{I}[p^{act}_t\ge\tau_{\mathrm{gate}}],0,1\right)$. A two-state hysteresis filter maps $\tilde{c}_t$ to discrete commit events: a commit is emitted when $\tilde{c}_t>\tau_{on}$ for $N_{dwell}$ consecutive windows; the state resets to \textsc{Hold} when $\tilde{c}_t<\tau_{off}$. Training proceeds in two stages: self-supervised pretraining (SSL) followed by supervised fine-tuning (FT). SSL optimizes a symmetric InfoNCE objective (maximizing agreement between two augmented views of the same window while separating different windows) with auxiliary masked reconstruction over two augmented views per window. FT minimizes a multi-objective loss over action, and task classification, and commit-readiness supervision, with sample-weighted reduction via quality weights $q_b$. Stochastic scenario sampling assigns each minibatch a sensor-availability scenario to reduce train--test mismatch. Fusion weights are regularized by entropy penalty $\mathcal{L}_{dom}=\mathrm{ReLU}(H_{\min}-H(\mathbf{w}))$ to prevent single-modality dominance. Validation calibrates thresholds ($\tau_{on}$, $\tau_{off}$, $N_{dwell}$, $N_{cool}$) to balance time-to-commit against false commits during \textsc{Rest}. Runtime was measured on an NVIDIA A100 80GB: inference latency is $7.89\pm0.08$\,ms per window at batch size 1, with 2.01M parameters.




\begin{table*}[t]
\centering
\scriptsize
\setlength{\tabcolsep}{2.8pt}
\renewcommand{\arraystretch}{1.15}
\caption{Action detection on LOSO across sensor dropout scenarios (S0--S6). Reported values are mean$\pm$std over folds. }
\label{tab:action_s0_s6_test_compact}
\resizebox{\textwidth}{!}{%
\begin{tabular}{l l c c c c c c c}
\toprule
\multirow{2}{*}{\textbf{Model}} & \multirow{2}{*}{\textbf{Metric}} &
\multicolumn{7}{c}{\textbf{Scenario}} \\
\cmidrule(lr){3-9}
&& \textbf{S0} & \textbf{S1} & \textbf{S2} & \textbf{S3} & \textbf{S4} & \textbf{S5} & \textbf{S6} \\
\midrule
\multirow{3}{*}{\textbf{Ours (NeuroCommitSSM)}}
& \oc{Action Bal-Acc ($\uparrow$)} 
& \oc{\textbf{0.950$\pm$0.022}} & \oc{\textbf{0.943$\pm$0.025}} & \oc{\textbf{0.854$\pm$0.045}} 
& \oc{\textbf{0.949$\pm$0.027}} & \oc{\textbf{0.785$\pm$0.075}} & \oc{\textbf{0.942$\pm$0.031}} & \oc{\textbf{0.701$\pm$0.040}} \\
& \oc{Action AUPRC ($\uparrow$)} 
& \oc{\textbf{0.993$\pm$0.005}} & \oc{\textbf{0.989$\pm$0.006}} & \oc{\textbf{0.967$\pm$0.016}} 
& \oc{\textbf{0.990$\pm$0.008}} & \oc{\textbf{0.927$\pm$0.042}} & \oc{\textbf{0.984$\pm$0.009}} & \oc{\textbf{0.828$\pm$0.043}} \\
& \oc{REST FPR ($\downarrow$)} 
& \oc{\textbf{0.052$\pm$0.043}} & \oc{\textbf{0.050$\pm$0.052}} & \oc{\textbf{0.172$\pm$0.100}} 
& \oc{\textbf{0.035$\pm$0.045}} & \oc{\textbf{0.219$\pm$0.151}} & \oc{\textbf{0.031$\pm$0.059}} & \oc{\textbf{0.514$\pm$0.086}} \\
\midrule


\multirow{3}{*}{MM-Transformer (Token Fusion)}
& Action Bal-Acc ($\uparrow$) & 0.887$\pm$0.073 & 0.886$\pm$0.071 & 0.606$\pm$0.056 & 0.829$\pm$0.094 & 0.508$\pm$0.022 & 0.828$\pm$0.094 & 0.600$\pm$0.057 \\
& Action AUPRC ($\uparrow$) & 0.967$\pm$0.048 & 0.967$\pm$0.048 & 0.760$\pm$0.076 & 0.962$\pm$0.059 & 0.660$\pm$0.088 & 0.962$\pm$0.059 & 0.760$\pm$0.065 \\
& REST FPR ($\downarrow$) & 0.112$\pm$0.162 & 0.113$\pm$0.161 & 0.307$\pm$0.200 & 0.247$\pm$0.212 & 0.685$\pm$0.438 & 0.247$\pm$0.213 & 0.716$\pm$0.216 \\
\midrule

\multirow{3}{*}{GRU (Gated Fusion)}
& Action Bal-Acc ($\uparrow$) & 0.898$\pm$0.067 & 0.765$\pm$0.128 & 0.652$\pm$0.096 & 0.893$\pm$0.078 & 0.619$\pm$0.092 & 0.754$\pm$0.145 & 0.567$\pm$0.047 \\
& Action AUPRC ($\uparrow$) & 0.980$\pm$0.021 & 0.941$\pm$0.048 & 0.826$\pm$0.091 & 0.981$\pm$0.018 & 0.838$\pm$0.089 & 0.951$\pm$0.028 & 0.761$\pm$0.077 \\
& REST FPR ($\downarrow$) & 0.124$\pm$0.140 & 0.041$\pm$0.073 & 0.595$\pm$0.272 & 0.134$\pm$0.162 & 0.665$\pm$0.279 & 0.042$\pm$0.102 & 0.77$\pm$0.153 \\
\midrule

\multirow{3}{*}{TCN (Early Concat Fusion)}
& Action Bal-Acc ($\uparrow$) & 0.863$\pm$0.080 & 0.793$\pm$0.091 & 0.610$\pm$0.046 & 0.830$\pm$0.116 & 0.556$\pm$0.044 & 0.742$\pm$0.113 & 0.623$\pm$0.061 \\
& Action AUPRC ($\uparrow$) & 0.971$\pm$0.023 & 0.949$\pm$0.032 & 0.914$\pm$0.044 & 0.964$\pm$0.034 & 0.856$\pm$0.077 & 0.921$\pm$0.059 & 0.821$\pm$0.050 \\
& REST FPR ($\downarrow$) & 0.158$\pm$0.189 & 0.052$\pm$0.067 & 0.749$\pm$0.109 & 0.210$\pm$0.273 & 0.860$\pm$0.102 & 0.092$\pm$0.135 & 1.64$\pm$0.204 \\
\midrule

\multirow{3}{*}{Mean Pool + Linear}
& Action Bal-Acc ($\uparrow$) & 0.782$\pm$0.084 & 0.782$\pm$0.083 & 0.517$\pm$0.041 & 0.776$\pm$0.099 & 0.500$\pm$0.001 & 0.776$\pm$0.099 & 0.517$\pm$0.041 \\
& Action AUPRC ($\uparrow$) & 0.921$\pm$0.043 & 0.922$\pm$0.042 & 0.773$\pm$0.067 & 0.916$\pm$0.041 & 0.626$\pm$0.077 & 0.917$\pm$0.041 & 0.774$\pm$0.070 \\
& REST FPR ($\downarrow$) & 0.287$\pm$0.207 & 0.286$\pm$0.206 & 0.931$\pm$0.150 & 0.322$\pm$0.218 & 1.000$\pm$0.001 & 0.322$\pm$0.219 & 0.931$\pm$0.149 \\
\bottomrule
\end{tabular}%
}
\end{table*}

\begin{table*}[t]
\centering
\scriptsize
\setlength{\tabcolsep}{2.8pt}
\renewcommand{\arraystretch}{1.15}
\caption{Task classification (LOSO) across sensor dropout scenarios (S0--S6). Reported values are mean$\pm$std over folds.}
\label{tab:task_s0_s6_test_compact}
\resizebox{\textwidth}{!}{%
\begin{tabular}{l l c c c c c c c}
\toprule
\multirow{2}{*}{\textbf{Model}} & \multirow{2}{*}{\textbf{Metric}} &
\multicolumn{7}{c}{\textbf{Scenario}} \\
\cmidrule(lr){3-9}
&& \textbf{S0} & \textbf{S1} & \textbf{S2} & \textbf{S3} & \textbf{S4} & \textbf{S5} & \textbf{S6} \\
\midrule
\multirow{2}{*}{\textbf{Ours (NeuroCommitSSM)}}
& \oc{Task Bal-Acc ($\uparrow$)} 
& \oc{\textbf{0.657$\pm$0.077}} & \oc{\textbf{0.630$\pm$0.088}} & \oc{\textbf{0.513$\pm$0.077}} 
& \oc{\textbf{0.544$\pm$0.104}} & \oc{\textbf{0.353$\pm$0.063}} & \oc{\textbf{0.505$\pm$0.129}} & \oc{\textbf{0.440$\pm$0.071}} \\
& \oc{Task AUPRC ($\uparrow$)} 
& \oc{\textbf{0.755$\pm$0.093}} & \oc{\textbf{0.720$\pm$0.104}} & \oc{\textbf{0.594$\pm$0.103}} 
& \oc{\textbf{0.621$\pm$0.120}} & \oc{\textbf{0.368$\pm$0.066}} & \oc{\textbf{0.564$\pm$0.150}} & \oc{\textbf{0.535$\pm$0.099}} \\
\midrule


\multirow{2}{*}{MM-Transformer (Token Fusion)}
& Task Bal-Acc ($\uparrow$) & 0.564$\pm$0.106 & 0.566$\pm$0.106 & 0.416$\pm$0.084 & 0.431$\pm$0.131 & 0.203$\pm$0.019 & 0.434$\pm$0.134 & 0.418$\pm$0.081 \\
& Task AUPRC ($\uparrow$) & 0.677$\pm$0.111 & 0.678$\pm$0.113 & 0.517$\pm$0.104 & 0.508$\pm$0.140 & 0.217$\pm$0.019 & 0.509$\pm$0.144 & 0.528$\pm$0.105 \\
\midrule

\multirow{2}{*}{GRU (Gated Fusion)}
& Task Bal-Acc ($\uparrow$) & 0.603$\pm$0.107 & 0.522$\pm$0.116 & 0.381$\pm$0.059 & 0.516$\pm$0.122 & 0.203$\pm$0.015 & 0.398$\pm$0.158 & 0.383$\pm$0.065 \\
& Task AUPRC ($\uparrow$) & 0.688$\pm$0.112 & 0.639$\pm$0.119 & 0.470$\pm$0.085 & 0.580$\pm$0.141 & 0.225$\pm$0.027 & 0.536$\pm$0.147 & 0.491$\pm$0.095 \\
\midrule

\multirow{2}{*}{TCN (Early Concat Fusion)}
& Task Bal-Acc ($\uparrow$) & 0.610$\pm$0.103 & 0.578$\pm$0.118 & 0.457$\pm$0.081 & 0.495$\pm$0.115 & 0.270$\pm$0.070 & 0.433$\pm$0.139 & 0.416$\pm$0.077 \\
& Task AUPRC ($\uparrow$) & 0.699$\pm$0.125 & 0.688$\pm$0.136 & 0.542$\pm$0.111 & 0.579$\pm$0.145 & 0.324$\pm$0.090 & 0.559$\pm$0.162 & 0.514$\pm$0.101 \\
\midrule

\multirow{2}{*}{Mean Pool + Linear}
& Task Bal-Acc ($\uparrow$) & 0.504$\pm$0.089 & 0.506$\pm$0.090 & 0.365$\pm$0.062 & 0.411$\pm$0.103 & 0.203$\pm$0.022 & 0.413$\pm$0.106 & 0.363$\pm$0.065 \\
& Task AUPRC ($\uparrow$) & 0.582$\pm$0.121 & 0.584$\pm$0.122 & 0.438$\pm$0.080 & 0.491$\pm$0.126 & 0.212$\pm$0.012 & 0.498$\pm$0.131 & 0.439$\pm$0.079 \\
\bottomrule
\end{tabular}%
}
\end{table*}

\subsection{System Integration and Hardware-in-the-Loop Validation}
\label{subsec:system_integration}

NeuroCommitSSM is integrated with a Kinova Gen3 through a ROS2 HAC supervisor, markerless RGB-D perception, and motion planning with IK, collision checking, and joint-limit validation. Perception estimates are transformed from the camera frame to the robot base frame using calibrated extrinsics and the ROS2 TF framework. The task-aware RGB-D pipeline provides a binary perception-feasibility cue $f_{\mathrm{cv}}(t)$ based on target visibility, depth support, scene stability, and obstacle proximity. Robot feasibility $f_{\mathrm{robot}}(t)$ checks IK solvability, collision-free planning, and joint-limit satisfaction. At each control update, HAC combines action-gated commit readiness with $f_{\mathrm{cv}}(t)$ and $f_{\mathrm{robot}}(t)$. The system enters \textsc{COMMIT} only when the dwell/hysteresis gate is active and both feasibility cues remain stable for a feasibility-dwell period; otherwise it remains in or returns to \textsc{HOLD} with controlled deceleration. During execution, perception is continuously re-evaluated, and motion is aborted if perception or robot feasibility becomes invalid. System-level validation is performed via repeatable HIL replay of 168 held-out tri-modal biosignal trials through the full stack without retuning the validation-calibrated decision parameters. Matched-condition comparisons across Commit-only, Feasibility-only, HOLD/COMMIT, and full HAC policies isolate the effect of feasibility-grounded shared autonomy. We report false starts, CV-infeasible starts, abort rate, success rate, time-to-success, and abort causes. HIL replay supports repeatable full-stack evaluation, but it does not replace live human-in-the-loop assistive testing with user adaptation, fatigue, trust, and closed-loop co-adaptation.

\section{Results}
\label{sec:results}

\begin{table*}[!htbp]
\centering
\scriptsize
\setlength{\tabcolsep}{1.75pt}
\renewcommand{\arraystretch}{1.12}
\caption{Commit decision quality across S0--S3 for FP/1k \textsc{Rest}, Flaps/min, and Commit Coverage (\%).}

\label{tab:commit_quality_test_s0_s3_tau1}
\resizebox{\textwidth}{!}{%
\begin{tabular}{l c c c c c c c c c c c c}
\toprule
\multirow{2}{*}{Method} & \multicolumn{3}{c}{S0} & \multicolumn{3}{c}{S1} & \multicolumn{3}{c}{S2} & \multicolumn{3}{c}{S3} \\
\cmidrule(lr){2-4}\cmidrule(lr){5-7}\cmidrule(lr){8-10}\cmidrule(lr){11-13}
& \makecell[c]{FP/1k\\rest} & \makecell[c]{Flaps\\ /min} & \makecell[c]{Commit Coverage (\%)} & \makecell[c]{FP/1k\\rest} & \makecell[c]{Flaps\\ /min} & \makecell[c]{Commit Coverage (\%)} & \makecell[c]{FP/1k\\rest} & \makecell[c]{Flaps\\ /min} & \makecell[c]{Commit Coverage (\%)} & \makecell[c]{FP/1k\\rest} & \makecell[c]{Flaps\\ /min} & \makecell[c]{Commit Coverage (\%)} \\
\midrule
\rowcolor{oursgray}
\cellcolor{white}Ours (NeuroCommitSSM) &
\textbf{0.75$\pm$0.97} & \textbf{6.46$\pm$3.45} & 30.06$\pm$19.27 &
\textbf{1.23$\pm$2.47} & \textbf{6.96$\pm$3.65} & 32.37$\pm$20.47 &
\textbf{0.12$\pm$0.64} & \textbf{3.06$\pm$1.52} & 14.01$\pm$6.86 &
\textbf{0.17$\pm$0.68} & \textbf{4.63$\pm$4.00} & 23.13$\pm$22.30 \\

MM-Transformer (Token Fusion) & 6.49$\pm$13.18 & 11.93$\pm$2.56 & 84.57$\pm$6.02 & 6.30$\pm$12.76 & 11.88$\pm$2.51 & 84.67$\pm$5.93 & 7.08$\pm$16.17 & 3.91$\pm$3.55 & 18.44$\pm$15.17 & 8.25$\pm$17.44 & 12.08$\pm$3.12 & 86.64$\pm$6.18 \\


TCN (Early Concat Fusion) & 8.56$\pm$13.23 & 11.62$\pm$2.55 & 83.32$\pm$8.72 & 2.44$\pm$3.44 & 10.00$\pm$3.10 & 60.41$\pm$19.48 & 80.15$\pm$30.62 & 18.10$\pm$5.83 & 85.71$\pm$8.36 & 7.28$\pm$14.44 & 11.43$\pm$2.53 & 81.03$\pm$10.23 \\
Mean Pool + Linear & 2.37$\pm$4.48 & 6.50$\pm$3.67 & 44.12$\pm$26.48 & 2.37$\pm$4.48 & 6.35$\pm$3.68 & 44.18$\pm$26.44 & 0.95$\pm$1.74 & 4.42$\pm$1.26 & 11.40$\pm$10.94 & 1.41$\pm$4.50 & 5.27$\pm$4.10 & 37.17$\pm$28.99 \\


Early-Fusion GRU (Hysteresis) & 9.44$\pm$15.43 & 8.51$\pm$2.66 & 59.41$\pm$15.42 & 4.75$\pm$8.56 & 7.31$\pm$2.34 & 42.19$\pm$17.92 & 86.01$\pm$25.17 & 18.75$\pm$5.63 & 77.58$\pm$11.02 & 20.99$\pm$21.49 & 10.62$\pm$2.94 & 66.55$\pm$15.77 \\


Unimodal EEG & 25.71$\pm$23.03 & 16.49$\pm$4.15 & 80.06$\pm$11.21 & -- & -- & -- & 25.71$\pm$23.03 & 16.49$\pm$4.15 & 80.06$\pm$11.21 & 25.71$\pm$23.03 & 16.49$\pm$4.15 & 80.06$\pm$11.21 \\
Unimodal EMG & 8.66$\pm$17.77 & 12.46$\pm$3.08 & 86.95$\pm$5.45 & 8.66$\pm$17.77 & 12.46$\pm$3.08 & 86.95$\pm$5.45 & -- & -- & -- & 8.66$\pm$17.77 & 12.46$\pm$3.08 & 86.95$\pm$5.45 \\
Unimodal Eye-Tracking & 8.95$\pm$5.75 & 6.47$\pm$1.75 & 39.72$\pm$10.35 & 8.95$\pm$5.75 & 6.47$\pm$1.75 & 39.72$\pm$10.35 & 8.95$\pm$5.75 & 6.47$\pm$1.75 & 39.72$\pm$10.35 & -- & -- & -- \\
\bottomrule
\end{tabular}%
}
\end{table*}

\begin{table*}[!htbp]
\centering
\scriptsize
\setlength{\tabcolsep}{1.90pt}
\renewcommand{\arraystretch}{1.12}
\caption{Commit decision quality across S4--S6 for FP/1k \textsc{Rest}, Flaps/min, and Commit Coverage (\%).}

\label{tab:commit_quality_test_s4_s6_tau1}
\resizebox{\textwidth}{!}{%
\begin{tabular}{l c c c c c c c c c}
\toprule
\multirow{2}{*}{Method} & \multicolumn{3}{c}{S4} & \multicolumn{3}{c}{S5} & \multicolumn{3}{c}{S6} \\
\cmidrule(lr){2-4}\cmidrule(lr){5-7}\cmidrule(lr){8-10}
& \makecell[c]{FP/1k\\rest} & \makecell[c]{Flaps\\ /min} & \makecell[c]{Commit Coverage (\%)} & \makecell[c]{FP/1k\\rest} & \makecell[c]{Flaps\\ /min} & \makecell[c]{Commit Coverage (\%)} & \makecell[c]{FP/1k\\rest} & \makecell[c]{Flaps\\ /min} & \makecell[c]{Commit Coverage (\%)} \\
\midrule
\rowcolor{oursgray}
\cellcolor{white}Ours (NeuroCommitSSM) & \textbf{0.29$\pm$0.81} & \textbf{0.41$\pm$0.77} & 17.7$\pm$3.71 & \textbf{0.89$\pm$2.88} & \textbf{4.54$\pm$4.29} & 22.76$\pm$24.16 & \textbf{0.35$\pm$0.71} & \textbf{3.32$\pm$1.81} & 15.36$\pm$8.06 \\


MM-Transformer (Token Fusion) & 11.12$\pm$37.28 & 2.09$\pm$6.78 & 7.27$\pm$23.02 & 7.89$\pm$16.61 & 12.09$\pm$3.04 & 86.68$\pm$6.14 & 6.72$\pm$16.75 & 3.81$\pm$3.65 & 18.04$\pm$15.49 \\


TCN (Early Concat Fusion) & 99.95$\pm$29.25 & 19.59$\pm$6.16 & 84.58$\pm$10.69 & 0.94$\pm$2.33 & 8.82$\pm$3.42 & 52.49$\pm$22.61 & 1.90$\pm$2.67 & 3.62$\pm$2.39 & 19.36$\pm$12.45 \\

Mean Pool + Linear & -- & -- & -- & 1.41$\pm$4.50 & 5.26$\pm$4.08 & 37.07$\pm$29.00 & 0.95$\pm$1.74 & 3.41$\pm$1.25 & 11.35$\pm$10.89 \\



Early-Fusion GRU (Hysteresis) & 125.90$\pm$22.99 & 22.61$\pm$6.08 & 88.82$\pm$9.84 & 15.71$\pm$16.53 & 7.95$\pm$2.59 & 46.62$\pm$19.43 & 27.44$\pm$30.66 & 6.93$\pm$6.01 & 31.43$\pm$22.50 \\


Unimodal EEG & 25.71$\pm$23.03 & 16.49$\pm$4.15 & 80.06$\pm$11.21 & -- & -- & -- & -- & -- & -- \\
Unimodal EMG & -- & -- & -- & 8.66$\pm$17.77 & 12.46$\pm$3.08 & 86.95$\pm$5.45 & -- & -- & -- \\
Unimodal Eye-Tracking & -- & -- & -- & -- & -- & -- & 8.95$\pm$5.75 & 6.47$\pm$1.75 & 39.72$\pm$10.35 \\
\bottomrule
\end{tabular}%
}
\end{table*}

NeuroCommitSSM is evaluated on action/task recognition, commit-event decision quality, and system-level execution under LOSO cross-validation with sensor-dropout scenarios. \textbf{LOSO is a stringent subject-independent protocol for biosignals, since EEG/EMG/ET are nonstationary and exhibit substantial inter-subject variability.} Metrics are reported as mean$\pm$std over folds, with paired Wilcoxon signed-rank tests and Holm correction. \textbf{Tables~\ref{tab:action_s0_s6_test_compact} and~\ref{tab:task_s0_s6_test_compact}} show that NeuroCommitSSM achieves the best action balanced accuracy in S0 (0.950$\pm$0.022) and remains competitive under sensor loss (S6: 0.701$\pm$0.040) while maintaining low REST false positives. \textbf{Tables~\ref{tab:commit_quality_test_s0_s3_tau1} and~\ref{tab:commit_quality_test_s4_s6_tau1}} report commit-event quality using the action-gated readiness stream $c_t^{\mathrm{eff}}$ (readiness gated by the action threshold) and a dwell/hysteresis commit filter that is reset at the start of each trial. To ensure fair commit-quality comparisons, all baselines use the same commit-event extraction as our method. FP/1k \textsc{Rest} counts false commit events (entries into \textsc{Commit}) per 1000 \textsc{Rest} windows, and flaps/min counts \textsc{Hold}$\leftrightarrow$\textsc{Commit} state changes (stride 0.25\,s). Commit Coverage (\%) is the fraction of trials with at least one commit event, so low FP rates are not explained by suppressing commits entirely. NeuroCommitSSM achieves the lowest false-commit rates across scenarios while retaining nonzero coverage under dropout; by contrast, threshold-based baselines can produce large rest false-commit rates under unimodal settings (e.g., TCN in S4: 99.95$\pm$29.25 FP/1k \textsc{Rest} vs.\ 0.29$\pm$0.81). Coverage is lower in S0 (${ \sim }30\%$) because conservative dwell/hysteresis gating intentionally rejects brief pre-onset and transient spikes to suppress \textsc{Rest} false commits. Task accuracy is lower under LOSO (S0: 0.657; S4: 0.353 vs.\ chance 0.20) due to cross-subject physiological variability and overlapping early task dynamics, which reduces separability when evaluating on unseen subjects. Execution is gated by commit readiness and HAC feasibility checks rather than task prediction, so initiation safety is unaffected. \textbf{Table~\ref{tab:ablation_s1s3avg_test}} summarizes ablations averaged over S1--S3. Removing scenario-mix training causes the largest drop in Macro-F1 and the largest increase in flaps/min, indicating that exposure to mixed sensor availability is critical for stable decisions. Removing feature injection produces the largest increase in FP/1k \textsc{Rest}, showing that compact engineered descriptors provide useful dropout cues. Other ablations (dominance regularization, uncertainty-aware fusion, reliability conditioning) yield smaller but consistent degradations. \textbf{Table~\ref{tab:comp_vs_baselines}} reports a min--max normalized composite over S0--S6 (Macro-F1, AUPRC, FP/1k \textsc{Rest}, flaps/min): NeuroCommitSSM improves over all baselines ($\tilde{\Delta}>0$), with the closest competitor (MM-Transformer Token Fusion) trailing by $\tilde{\Delta}=0.177$ ($p_{\mathrm{Holm}}^{\max}=3.5\times10^{-8}$); all Holm-corrected Wilcoxon tests remain significant ($\le 5.3\times10^{-6}$).

\begin{table}[!htbp]
\centering
\scriptsize
\setlength{\tabcolsep}{2.1pt}
\renewcommand{\arraystretch}{1.10}
\caption{Ablation study (averaged over S1--S3).}


\label{tab:ablation_s1s3avg_test}
\resizebox{\columnwidth}{!}{%
\begin{tabular}{l c c c c}
\toprule
\textbf{Variant} &
\makecell{\textbf{Action}\\\textbf{Macro-F1} $\uparrow$} &
\makecell{\textbf{FP/1k}\\\textbf{REST} $\downarrow$} &
\makecell{\textbf{Flaps}\\\textbf{/min} $\downarrow$} &
\makecell{\textbf{Action}\\\textbf{AUPRC} $\uparrow$} \\
\midrule
Base & $0.914\pm0.036$ & $7.481\pm5.056$ & $12.283\pm3.295$ & $0.981\pm0.009$ \\
No scenario-mix training & $0.887\pm0.037$ & $7.373\pm5.248$ & $15.153\pm3.295$ & $0.968\pm0.012$ \\
No dominance regularization & $0.905\pm0.035$ & $7.748\pm5.629$ & $12.579\pm3.367$ & $0.978\pm0.009$ \\
No uncertainty-aware fusion & $0.907\pm0.035$ & $7.442\pm5.758$ & $13.763\pm3.425$ & $0.979\pm0.009$ \\
No reliability conditioning & $0.907\pm0.034$ & $7.450\pm4.925$ & $12.699\pm2.957$ & $0.979\pm0.008$ \\
No feature injection & $0.896\pm0.038$ & $8.508\pm6.170$ & $13.580\pm2.894$ & $0.977\pm0.010$ \\

\bottomrule
\end{tabular}%
}
\vspace{1mm}

\end{table}


\begin{table}[!t]
\centering
\scriptsize
\setlength{\tabcolsep}{2.5pt}
\renewcommand{\arraystretch}{1.05}
\caption{Composite decision-quality vs.\ baselines. $\tilde{\Delta}$: median improvement over S0--S6 on a min--max normalized composite (Macro-F1, AUPRC, FP/1k REST, flaps/min).} 


\label{tab:comp_vs_baselines}
\begin{tabular}{@{}p{0.46\columnwidth}ccc@{}}
\toprule
Baseline & $\tilde{\Delta}$ & $p_{\mathrm{Holm}}^{\max}$ & Mean rank \\
\midrule
Unimodal EMG & 0.172 & 1.2e-07 & 2.57 \\
MM-Transformer (Token Fusion) & 0.177 & 3.5e-08 & 2.43 \\
Mean Pool + Linear & 0.205 & 1.2e-07 & 4.57 \\
TCN (Early Concat Fusion) & 0.218 & 5.7e-07 & 6.14 \\
Early-Fusion GRU (Hysteresis) & 0.247 & 5.3e-08 & 8.57 \\
Unimodal EEG & 0.272 & 7.8e-08 & 8.00 \\
Unimodal Eye-Tracking & 0.332 & 1.3e-08 & 9.29 \\
\bottomrule
\end{tabular}
\end{table}


\textbf{Table~\ref{tab:rq2_hac_ablation_cv}} reports HIL evaluation of the ROS2 supervisor on a Kinova Gen3 using repeatable biosignal replay, totaling 168 condition-runs (42 paired physical trials replayed under 4 supervisor variants) across the ADL tasks. Here, \emph{false starts} denote motion initiations during ground-truth \textsc{Rest}; \emph{CV-infeasible starts} denote initiations when perception feasibility indicated an invalid/unstable target estimate; \emph{abort rate} is the fraction of trials with outcome \textsc{Abort}; \emph{success rate} is the fraction completing the intended behavior; and \emph{time-to-success} is reported as the mean over all trials with non-success trials assigned 0~s (thus increasing when more trials reach completion). Full HAC yields the lowest false starts and abort rate and the highest success rate, while ablating ASSIST or feasibility-stability gating degrades both safety and completion. We report 95\% Wilson CIs for all binomial rates. Because each physical trial is replayed under all supervisors ($n{=}42$ paired trials), success comparisons are paired; we apply Cochran's $Q$ across supervisor conditions ($Q{=}49.61$, dof$=3$, $p{=}9.67{\times}10^{-11}$), followed by exact McNemar post-hoc tests with Holm correction (e.g., Full HAC vs Commit-only: $p_{\text{Holm}}{=}7.15{\times}10^{-7}$; vs Feasibility-only: $p_{\text{Holm}}{=}7.75{\times}10^{-6}$; vs HOLD/COMMIT: $p_{\text{Holm}}{=}1.27{\times}10^{-2}$).


\begin{table*}[!htbp]
\centering
\caption{HIL supervisor ablations on a Kinova Gen3 arm using commit-readiness $c(t)$ and CV-based feasibility $f_{\text{cv}}(t)$.}
\label{tab:rq2_hac_ablation_cv}
\small
\setlength{\tabcolsep}{3.5pt}
\renewcommand{\arraystretch}{1.1}
\begin{tabular}{@{}l p{4.2cm} c c c c c@{}}
\toprule
\textbf{Supervisor variant} &
\textbf{Start rule (initiation)} &
\makecell{\textbf{False}\\\textbf{starts}$^a$ $\downarrow$} &
\makecell{\textbf{CV-infeas.}\\\textbf{starts}$^b$ $\downarrow$} &
\makecell{\textbf{Abort}\\\textbf{rate}$^c$ $\downarrow$} &
\makecell{\textbf{Success}\\\textbf{rate} $\uparrow$} &
\makecell{\textbf{Time-to-}\\\textbf{success (s)}} \\
\midrule
Commit-only (baseline) &
Start on $\texttt{gate}(c)=1$ &
11.9\% & 100.0\% & 0.0\% & 40.5\% & 16.8 \\
Feasibility-only (CV) &
Start on $\texttt{gate}\!\big(f_{\text{cv}}\big)=1$ &
100.0\% & 0.0\% & 54.8\% & 42.9\% & 19.4 \\
HOLD/COMMIT (no ASSIST) &
Start on $\texttt{gate}(c)=1 \wedge f_{\text{cv}}=1$ &
7.1\% & 0.0\% & 19.0\% & 73.8\% & 33.8 \\
\rowcolor{oursgray}
\cellcolor{white}\textbf{Full HOLD--ASSIST--COMMIT} &
Start if $\texttt{gate}(c)\wedge \texttt{stable}(f_{\text{cv}})$ (Assist $\rightarrow$ feasible) &
\textbf{2.4\%} & \textbf{0.0\%} & \textbf{2.4\%} & \textbf{97.6\%} & \textbf{46.3} \\
\bottomrule
\end{tabular}
\vspace{1mm}
\end{table*}

\section{Conclusion and Future Work}

NeuroCommitSSM predicts commit readiness from synchronized EEG--EMG--ET and combines dwell/hysteresis filtering with feasibility-gated HOLD--ASSIST--COMMIT control, reducing \textsc{Rest} false activations and decision instability under LOSO and sensor dropout while providing early-stage system-level validation through Kinova Gen3 HIL replay. Future work will evaluate the pipeline with individuals with motor impairments or physical disabilities, incorporate standardized usability and workload assessments, and explore transfer learning and subject-specific adaptation to better handle inter-user variability, weak or missing EMG, and real-world closed-loop deployment. We will also study embedded deployment on edge devices and optimize the safety--responsiveness trade-off using adaptive dwell/hysteresis conditioned on signal quality and task phase while preserving low \textsc{Rest} false commits.




\bibliographystyle{ieeetr}
\bibliography{refs}

\end{document}